\documentclass[sigconf,natbib=true,anonymous=false]{acmart}
\renewcommand\footnotetextcopyrightpermission[1]{} 
\AtBeginDocument{%
  }

\acmConference[Conference]{2025}{Under Review}{}

\makeatletter
\gdef\@copyrightpermission{
}
\makeatother

\usepackage[most]{tcolorbox}

\usepackage[normalem]{ulem}
\usepackage{tabularray}
\usepackage{makecell} 
\usepackage[normalem]{ulem}
\useunder{\uline}{\ul}{}
\usepackage{algorithm}
\usepackage{algorithmic}
\usepackage{multirow}
\usepackage{amsmath} 
\usepackage{enumitem}
\usepackage{balance}
\usepackage{caption} 

\usepackage{array} 
\usepackage{etoolbox} 
\usepackage{graphicx}
\usepackage{booktabs}

\begin{document}
\title{Parametric Retrieval Augmented Generation}

\author{Weihang Su}
\email{swh22@mails.tsinghua.edu.cn}
\affiliation{
DCST, Tsinghua University\\
Beijing 100084 \country{China}
}

\author{Yichen Tang}
\authornote{Contributed equally}
\affiliation{
DCST, Tsinghua University\\
Beijing 100084 \country{China}
}

\author{Qingyao Ai}
\authornote{Corresponding author}
\email{aiqy@tsinghua.edu.cn}
\affiliation{
DCST, Tsinghua University\\
Beijing 100084 \country{China}
}

\author{Junxi Yan}
\affiliation{
DCST, Tsinghua University\\
Beijing 100084 \country{China}
}

\author{Changyue Wang}
\affiliation{
DCST, Tsinghua University\\
Beijing 100084 \country{China}
}

\author{Hongning Wang}
\affiliation{
DCST, Tsinghua University\\
Beijing 100084 \country{China}
}

\author{Ziyi Ye}
\affiliation{
DCST, Tsinghua University\\
Beijing 100084 \country{China}
}

\author{Yujia Zhou}
\affiliation{
DCST, Tsinghua University\\
Beijing 100084 \country{China}
}

\author{Yiqun Liu}
\affiliation{
DCST, Tsinghua University\\
Beijing 100084 \country{China}
}

\renewcommand{\shortauthors}{Su, et al.}

\begin{abstract}

Retrieval-augmented generation (RAG) techniques have emerged as a promising solution to enhance the reliability of large language models (LLMs) by addressing issues like hallucinations, outdated knowledge, and domain adaptation.
In particular, existing RAG methods append relevant documents retrieved from external corpus or databases to the input of LLMs to guide their generation process, which we refer to as the in-context knowledge injection method. 
While this approach is simple and often effective, it has inherent limitations. Firstly, increasing the context length and number of relevant documents can lead to higher computational overhead and degraded performance, especially in complex reasoning tasks. More importantly, in-context knowledge injection operates primarily at the input level, but LLMs store their internal knowledge in their parameters. This gap fundamentally limits the capacity of in-context methods. To this end, we introduce Parametric retrieval-augmented generation (Parametric RAG), a new RAG paradigm that integrates external knowledge directly into the parameters of feed-forward networks (FFN) of an LLM through document parameterization. This approach not only saves online computational costs by eliminating the need to inject multiple documents into the LLMs' input context, but also deepens the integration of external knowledge into the parametric knowledge space of the LLM. Experimental results demonstrate that Parametric RAG substantially enhances both the effectiveness and efficiency of knowledge augmentation in LLMs. Also, it can be combined with in-context RAG methods to achieve even better performance\footnote{We have open-sourced all the code, data, and models in the following anonymized GitHub link: https://github.com/oneal2000/PRAG}.

\end{abstract}

\keywords{Large Language Model, Retrieval Augmented Generation, Knowledge Representation, Parametric Information Representation}

\maketitle

\section{Introduction}
\label{sec-intro}

\begin{figure*}[t]
\centering
    \includegraphics[width=\textwidth]{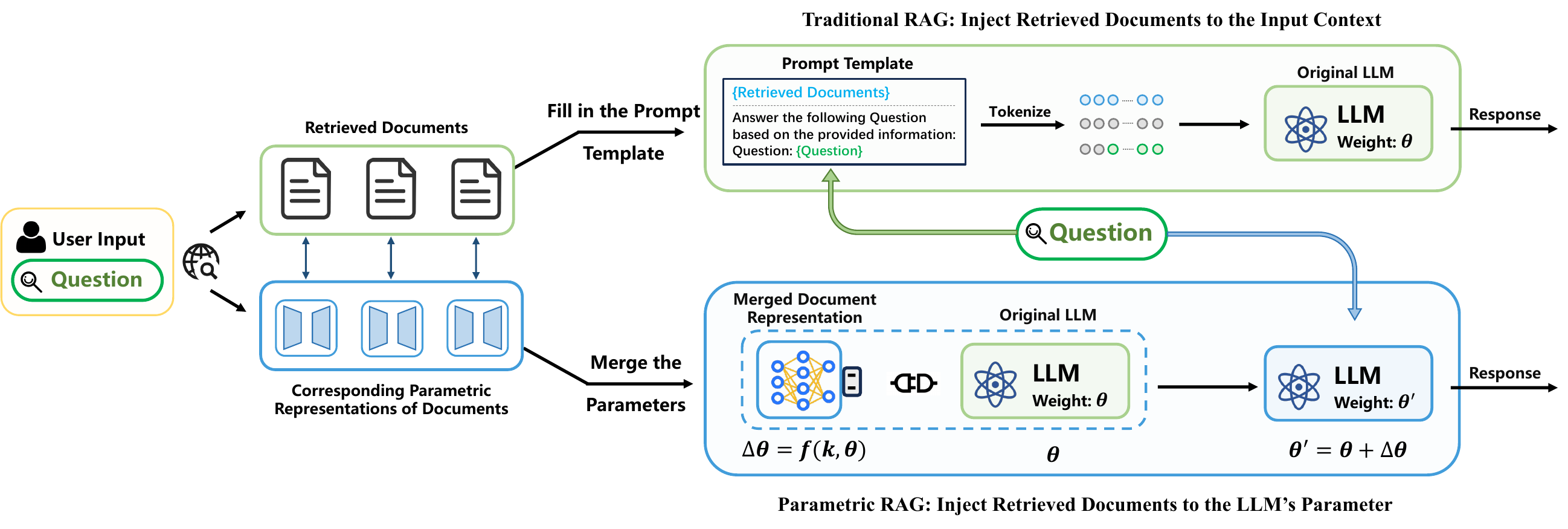}
    \caption{An illustration of the comparison of in-context RAG and our proposed Parametric RAG paradigms: In-context RAG combines the tokens of relevant documents and the query in the input, using the original LLM $\theta$ to answer the question without modifying its parameters. Our proposed Parametric RAG updates the LLM’s parameters $\theta^{\prime} = \theta + \Delta \theta$ based on the retrieved documents, \textit{temporarily} integrating relevant knowledge into LLM's parameters to answer the question.} 
    \label{pic:framework}
\end{figure*}

Large Language Models (LLMs) have demonstrated remarkable capabilities across a wide range of information retrieval (IR) and natural language processing (NLP) tasks~\cite{brown2020language,fang2024scaling,chowdhery2022palm,touvron2023llama,scao2022bloom,yang2024qwen2}. 
Despite these successes, a critical limitation remains: once training is complete, an LLM's internal knowledge becomes effectively static, making it challenging to incorporate newly emerging information or knowledge not included in its pre-training data.
To address this challenge, retrieval-augmented generation (RAG) has emerged as a prominent solution.
RAG enables LLMs to dynamically access and utilize information beyond their pre-trained parameters by retrieving relevant information from an external corpus, thus improving their adaptability and performance~\cite{borgeaud2022improving,su2024dragin,lewis2020retrieval,su2024mitigating,guu2020retrieval,izacard2020leveraging,jiang2022retrieval,su2024unsupervised}.

Existing studies have explored various aspects of the RAG pipeline, considering factors such as retrieval timing~\cite{asaiself,su2024dragin,jiang2022retrieval,su2024mitigating}, document selection~\cite{yu2024rankrag,ke2024bridging}, and external knowledge organization~\cite{edge2024local,hu2024grag,peng2024graph}. 
While these innovations improve different stages of the pipeline, all RAG methods, regardless of their variations, share a common characteristic: they inject external knowledge by directly adding passages or documents into the input context of LLMs, which we refer to as the in-context knowledge injection.

Although this in-context knowledge injection approach is straightforward and often effective, recent studies have highlighted several limitations of this paradigm. 
First, injecting knowledge through input prompts will inevitably increase the context length. Long context not only introduces extra computational overhead and latency for LLM inference, but also hurts the performance of LLMs in understanding and utilizing external knowledge, especially in tasks that involve complex reasoning~\cite{liu2024lost,levy2024same}. 
Second, more importantly, the way LLMs process information in context is fundamentally different from the way they utilize internal knowledge stored in their parameters. 
Studies have shown that LLMs store most of their knowledge within the parameters of their neural network architecture (e.g., the parameters of their feed-forward network layers)~\cite{yu-ananiadou-2024-neuron,nanda2023fact}. 
Adding passages or documents in the input context could only affect the online computation of key-value (KV) pairs in the attention networks of LLMs, but not the model’s stored parameters, where its knowledge is encoded~\cite{yu-ananiadou-2024-neuron}.
This means that LLMs may never be able to utilize external knowledge as effectively as they use their internal knowledge in in-context RAG methods.  
A straightforward solution to this problem is to conduct supervised fine-tuning (SFT) with retrieved documents, thereby incorporating relevant knowledge directly into the LLM’s parameters. However, such SFT-based methods are considered suboptimal as they require substantial computational resources and GPU memory, making it impractical to inject relevant documents online for every query. 
In addition, they can negatively affect the original ability of the LLM to follow instructions~\cite{wu2024continual,dong2023abilities} and lack the flexibility of in-context methods, which allow external knowledge to be added or removed on the fly.

The observations above inspire us to raise the following research question:
\textit{Is it possible to {inject external knowledge into LLM parameters }effectively, efficiently, and flexibly for retrieval-augmented generation?}

To this end, we introduce a new RAG paradigm, \textbf{Parametric Retrieval Augmented Generation (Parametric RAG)}, which directly injects the external knowledge into the Feed-Forward Networks (FFN) of an LLM.
Specifically, our approach begins with an offline preprocessing phase that parameterizes each document from the external corpus, transforming them into a small set of parameters (usually a few MB per document) that can be directly integrated into the downstream LLM. We refer to this set of parameters as the parametric representation of the document. 
In the inference phase, we conduct retrieval augmented generation following a Retrieve-Update-Generate (RUG) workflow as shown in Figure~\ref{pic:framework}. 
The Retrieve step retrieves top-n documents from the external corpus based on the input prompt following the same procedure used by the existing RAG pipeline. Then, the Update step uses the parametric representations of the retrieved documents to update the LLM. Finally, in the Generate step, we use the updated LLMs to conduct inference directly based on the original input prompt.

Theoretical and empirical analysis show that our Parametric RAG method has superior inference efficiency and outperforms state-of-the-art in-context methods on several RAG benchmarks that involve tasks with complex reasoning. 
While the preprocessing phase of Parametric RAG introduces an offline computational overhead, this cost is affordable and even neglectable compared to the online cost of large-scale inference requests, leading to long-term savings in terms of power and carbon footprints.
Also, similar to in-context RAG, our method can adapt to various numbers of input documents on the fly. Furthermore, our proposed Parametric RAG pipeline is in parallel with existing in-context methods. As shown in our experiments, combining our methods with in-context RAG could produce even better performance on the benchmark datasets. This indicates that parametric knowledge injection could be a fruitful direction for the future development of the RAG system.

In summary, this paper makes the following key contributions:
\vspace{-0.5mm}

\begin{itemize}[leftmargin=*]

\item We propose Parametric RAG, a new RAG paradigm that integrates external knowledge directly into LLM's parameters.

\item We propose an offline method to build parametric document representation and a Retrieve-Update-Generate pipeline to conduct Parametric RAG on the fly.

\item We conduct extensive experiments to compare the state-of-the-art in-context RAG methods with our method to demonstrate the potential of Parametric RAG in terms of effectiveness and efficiency.

\end{itemize}

\section{Related Work}
\label{sec-related}

Large language models have shown remarkable performance across diverse applications. However, their inherent knowledge often proves insufficient for tackling knowledge-intensive tasks, underscoring the necessity of integrating external knowledge for robust performance in these contexts.
One prominent approach to address this gap is Retrieval-Augmented Generation (RAG), which improves LLMs by integrating relevant external knowledge sources~\cite{borgeaud2022improving,lewis2020retrieval,wang2024knowledge,su2024stard,wang2024lekube,guu2020retrieval,izacard2020leveraging,dong2025decoupling,jiang2022retrieval,shi2023replug}. 
In the traditional RAG framework, an external retriever~\cite{zhai2008statistical,su2023caseformer,robertson2009probabilistic,su2023wikiformer,ma2023caseencoder,dong2023i3,su2023thuir2} or a more complex retrieval system~\cite{salemi2024towards,li2023towards} retrieves relevant documents based on a query. 
These documents are then appended to the LLM's input context, allowing the LLM to leverage knowledge beyond its training data~\cite{lewis2020retrieval}.

Building upon the traditional RAG framework, several extensions have been proposed to improve its effectiveness and efficiency. 
One such extension, Adaptive RAG~\cite{jeong2024adaptive,wang2023self}, introduces adaptive retrieval strategies that actively adjust the retrieval pipeline based on the complexity of the query to improve the adaptability of RAG in different tasks. 
From another angle, to make in-context knowledge injection more effective in RAG scenarios, IR-CoT~\cite{trivedi2022interleaving} designs prompt templates specifically tailored for RAG that demonstrate how to perform chain-of-thought (CoT) reasoning based on the given passage. 
Each sentence in the CoT reasoning content is then applied to retrieve more relevant documents.
Another research direction, GraphRAG~\cite{edge2024local,hu2024grag,peng2024graph}, employs pre-constructed knowledge graphs to retrieve graph elements that encapsulate relational knowledge relevant to the query. 
GraphRAG has demonstrated enhanced performance, particularly in tasks that require structured, relational information.
In the context of long-form generation, where the LLM’s informational needs may change during the generation process, dynamic RAG techniques have been developed to actively decide when and what to retrieve during the generation process~\cite{su2024dragin,jiang2023active,su2024mitigating,liu2024ctrla,yao2024seakr,baek2024probing}. 
For example, FLARE~\cite{jiang2023active} triggers the retrieval module when the model’s token prediction probability falls below a predefined threshold. 
Similarly, DRAGIN~\cite{su2024dragin} models the real-time information needs of the LLM, generating queries based on the model’s internal state and preceding context to fetch relevant external knowledge dynamically.

To summarize, existing RAG approaches have explored various aspects of the RAG pipeline, considering factors such as retrieval timing~\cite{su2024dragin,jiang2023active,su2024mitigating,liu2024ctrla,yao2024seakr,baek2024probing}, prompt template for in-context knowledge injection~\cite{wang2024rat,trivedi2022interleaving}, document selection~\cite{yu2024rankrag}, and external knowledge organization\cite{edge2024local,hu2024grag,peng2024graph}. 
While these innovations improve different stages of the pipeline, all RAG methods, regardless of their variations, share a common characteristic at the knowledge injection level: relevant passages or documents are appended directly to the LLM’s input context to inject external knowledge.
In contrast, our proposed Parametric RAG paradigm diverges from all the existing RAG frameworks by directly injecting documents into the LLM’s parameters. 
This shift in knowledge integration addresses the inherent limitations of the in-context knowledge injection methods employed in all existing RAG frameworks.

\section{Methodology}
\label{sec:method}
In this section, we introduce our proposed {Parametric RAG} framework, shown in Figure ~\ref{pic:framework}.
This section begins by formulating the problem and providing an overview of the Parametric RAG framework (\S 3.1). 
Next, we introduce the Offline Document Parameterization process (\S\ref{sec:knowledge}), which transforms documents into parametric representations through \textit{\textbf{Document Augmentation}} and \textit{\textbf{Parametric Document Encoding}}. 
Finally, we introduce the Online Inference procedure (\S\ref{sec:inference}), where the parametric representations are retrieved, merged, and integrated into the LLM to generate responses.

\subsection{Problem Formulation and Overview}

This subsection introduces the problem formulation of the RAG task and provides an overview of our proposed Parametric RAG pipeline. 
Consider an LLM (denoted as \(\mathcal L\)) with base parameters \(\theta\). Given a user query \(q\), we aim to generate an accurate response using an external corpus \(K\). 
Formally, the corpus \(K\) is defined as:
$K = \{d_1, d_2, \ldots, d_N\}$,
where each \(d_i\) represents a text chunk, such as documents, Wikipedia articles, or passages (for convenience, we refer to each $d_i$ as ‘document’ in the following sections). 
The system contains a retrieval module \(R\) that calculates the relevance score of each document \(\{S_{d_1}, S_{d_2}, \ldots, S_{d_N}\}\) corresponding to the query \(q\).
Traditional RAG paradigms select the top \(k\) documents with the highest relevance scores as relevant external knowledge and append them to the input context of the \(\mathcal L\). 
This process is typically guided by a prompt template that instructs \(\mathcal L\) to generate the response based on the provided knowledge.

In contrast to the in-context RAG paradigm that injects relevant documents into the LLM's input context, 
in Parametric RAG, we propose to insert documents directly into the parameters of \(\mathcal L\).
To achieve this, the Parametric RAG framework is designed with two stages: an offline document parameterization stage and an online inference stage with a Retrieve-Update-Generate workflow.

\begin{figure*}[t]
\centering
    \includegraphics[width=0.99\textwidth]{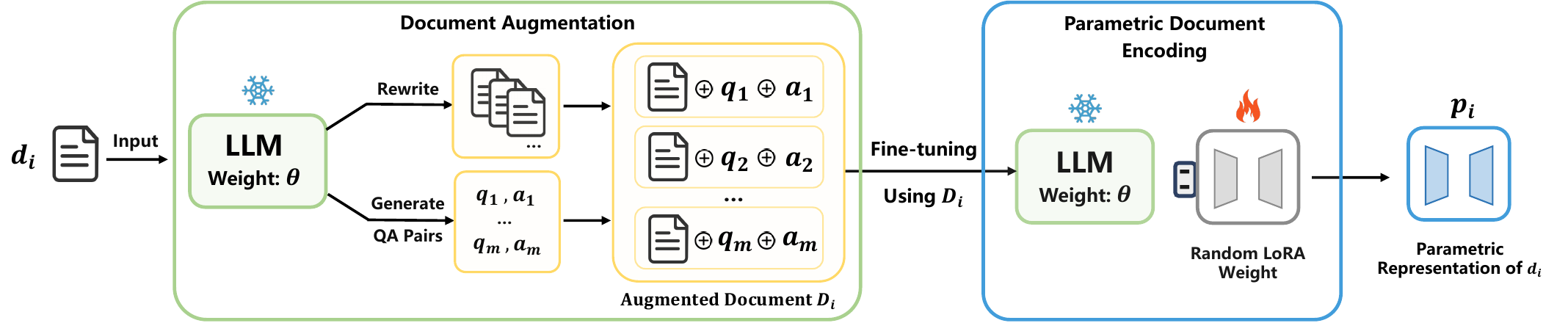}
\caption{An illustration of how we parameterize each document $d_i$ in the corpus during the \textit{Offline Document Parameterization} stage.}
\label{fig:method}
\end{figure*}

\paragraph{Offline document Parameterization} In this step (illustrated in Figure ~\ref{fig:method}), we offline transform each document in \(K\) into a parametric representation, thereby forming a set of parameters known as the Parametric Corpus \(K_P\). Specifically, we define:

\begin{equation}
K_P = \{ p_i \mid p_i = f_{\phi}(d_i), \quad i = 1, 2, \ldots, N \},
\label{eq:parametric_kb}
\end{equation}

\noindent where \(f_{\phi}\) is a mapping function that converts each document \(d_i\) into its corresponding parametric representation \(p_i\). We define parametric representations \(p_i\) to possess the following properties:

\begin{enumerate}[leftmargin=*]
    \item The parameters \(p_i\) can be plugged into the feed-forward network weights of the LLM.
    
    \item After inserting the parametric representation \(p_i\) into \(L\), the LLM can effectively comprehend the knowledge contained within the corresponding document \(d_i\).
    
    \item Different document parameters \(p_i\) can be merged through specific algorithms, and after merging, the LLM can grasp the combined knowledge corresponding to the merged documents.
\end{enumerate}

\paragraph{Online Inference} 
During the online inference process, our method first merges the parametric representations corresponding to the retrieved top-k documents and then plugs the merged parameters into the LLM. 
Subsequently, the updated LLM is used to answer the user’s question.
This overall framework allows for more efficient and effective knowledge injection, overcoming the limitations of traditional RAG by leveraging parameterized representations of external knowledge.

\subsection{Offline Document Parameterization}
\label{sec:knowledge}

In this subsection, we describe the detailed process of offline parameterizing each document in the corpus $K$ during the pre-processing phase.
Given a document $d_i$ and an LLM $\mathcal{L}$, our objective is to construct a parametric representation $p_i$ of the document. 
This representation enables $\mathcal{L}$ to effectively comprehend and utilize the knowledge contained in $d_i$ during inference. 
To achieve this, we propose a two-step approach: \textbf{\textit{Document Augmentation}} and \textbf{\textit{Parametric Document Encoding}}. These steps are combined to generate robust and informative parametric representations for each document.

\subsubsection{Document Augmentation}
\label{sec:DA}

Existing studies have shown that effectively incorporating factual knowledge from external documents into LLMs requires more than simply pre-training the model on raw text via standard next-token prediction. 
For example, \citeauthor{allen2023physics} \cite{allen2023physics} find that after being trained repeatedly on a given document, LLMs can memorize its content but fail to extract and apply this knowledge effectively in downstream tasks such as question answering. 
To address this issue, \citeauthor{allen2023physics} \cite{allen2023physics} propose two key strategies: (1) incorporating question-answer (QA) pairs derived from the document during training and (2) augmenting the document through multiple rewrites that express the same factual content in different forms. 
Their findings indicate that these two approaches enable LLMs to internalize knowledge to support accurate application in downstream tasks rather than reproducing the original text token-by-token\footnote{Our experimental results corroborate these conclusions, as shown in Figure~\ref{pic:exp}.}.

Building upon the insights discussed above, we introduce the document augmentation process consisting of two steps: \textbf{Document Rewriting} and \textbf{QA Pair Generation}, to construct robust and informative parametric representations for documents.
Specifically, for each document, we prompt\footnote{Due to space constraints, we have not included the specific prompts in the main text. However, all prompts used in this study are available at the following anonymous link: https://github.com/oneal2000/PRAG/blob/main/all\_prompt.md} the LLM $\mathcal L$ to {rewrite} the content multiple times using different wording, styles, or organizational structures. 
Formally, each document \( d_i \) is transformed into multiple rewritten documents \( \{ {d_i}^1, {d_i}^2, \dots, {d_i}^n \} \), which preserve the original facts but vary in language expression. 
Once each document has been rewritten into multiple documents, we prompt $\mathcal{L}$ again to generate {question-answer (QA) pairs} based on the original document \( d_i \). 
For each document \( d_i \), $\mathcal{L}$ produces a set of questions and their corresponding answers:
$\{({q_i}^1, {a_i}^1), ({q_i}^2, {a_i}^2), \dots, ({q_i}^m, {a_i}^m)\}$,
where \(m\) is a tunable hyperparameter representing the number of QA pairs we aim to generate per document. 
Integrating multiple rewrites with corresponding QA pairs transforms each document $d_i$ into a more comprehensive resource $D_i$ that preserves its original factual content while incorporating diverse linguistic variations. 
Formally:

\begin{equation}
\label{eq:1}
{D_i} = \{({d_i}^k, {{q_i}^j}, {{a_i}^j})\mid 1 \le k \le n,\, 1 \le j \le m\},
\end{equation}

\noindent where each \(({d_i}^k, {q_i}^j, {a_i}^j)\) triple corresponds to a rewritten document \({d_i}^k\) from the original document \(d_i\), coupled with a question \({q_i}^j\) and its respective answer \({a_i}^j\).

\subsubsection{Parametric Document Encoding}
\label{sec:additional-param-training}

In this subsection, we introduce the second step of the offline document parameterization pipeline, where we leverage the augmented dataset $D_i$ (defined in Eq.~\ref{eq:1}) to train the parametric representation $p_i$ for each document $d_i$. 
Specifically, we initialize these parametric representations as low-rank matrices corresponding to the feed-forward network (FFN) parameter matrix $W$ of the LLM $\mathcal{L}$, following the LoRA approach~\cite{hulora}.
This design allows each document \( d_i \) to be associated with independently trained low-rank parameters, allowing the model to internalize the knowledge specific to $d_i$ in a parameter-efficient manner.\footnote{Other parameter-efficient methods (e.g., Adapters or Prefix-Tuning) could also be used; exploring them is left for future work. In this work, we chose LoRA because it offers practical advantages over other alternatives. For example, LoRA is easier to merge compared to Adapters, and it requires less computational overhead during inference compared to Prefix-Tuning.}

Suppose the Transformer layers in \(\mathcal{L}\) have a hidden dimension \(h\), and the feed-forward network (FFN) within each layer has an intermediate dimension \(k\). Consequently, each FFN layer of \(\mathcal{L}\) contains a weight matrix \(W \in \mathbb{R}^{h \times k}\).
To incorporate document-specific knowledge, we introduce low-rank matrices \( A \) and \( B \) such that
\begin{align}
    W' \;=\; W + \Delta W \;=\; W +  A \, B^\top ,
    \label{eq:lora_app}
\end{align}
where \( A \in \mathbb{R}^{h \times r} \) and \( B \in \mathbb{R}^{k \times r} \), with \( r \ll \min(h, k) \).
The original weight matrix \(W\) is kept fixed, while \(A\) and \(B\) are the only trainable parameters for that layer. 
We denote these newly introduced parameters as \(\Delta \theta = \{A, B\}\). 
By combining the pre-trained weights \(W\) and \(\Delta \theta\), the model obtains the knowledge from the selected document. 
Each document in the corpus is associated with its instance of \(\Delta \theta\). 

For each document \(d_i\), we train its corresponding parametric representation \(\Delta \theta\) using its corresponding augmented dataset \({D}_i\). 
Recall from Eq.~\ref{eq:1} that \({D}_i\) contains triplets \(\bigl({d_i}^k, {q_i}^j, {a_i}^j\bigr)\).
For each triplet, we concatenate \({d_i}^k\), \({q_i}^j\), and \({a_i}^j\) into a token sequence:
\begin{align}
    x \;=\; [\,{d_i}^k \oplus \, {q_i}^j  \oplus \, {a_i}^j\,],
\end{align}
where \([\,\cdot\,\oplus\,\cdot\,]\) indicates concatenation. Let \(T\) be the total number of tokens in \(x\).
We adopt a standard sequential language modeling objective to ensure that the LLM internalizes knowledge from the entire augmented text (i.e., both the documents and QA pairs). Specifically, we optimize:
\begin{align}
    \label{eq:lm_training}
    \min_{\Delta \theta} \sum_{({d_i}^k, {q_i}^j, {a_i}^j) \in {D}_i} 
    \sum_{t=1}^{T} 
    -\log \; P_{\theta + \Delta \theta}\bigl(x_t \,\big|\, x_{<t}\bigr),
\end{align}
where \(\theta\) are the frozen pretrained parameters of the LLM, and \(\Delta \theta = \{A, B\}\) are the trainable low-rank matrices introduced in Eq.~\ref{eq:lora_app}. 
The innermost summation is taken over all tokens \(x_t\) in the concatenated input sequence (document, question, and answer)\footnote{The loss is computed not only on the answer but also across the entire concatenated sequence, including the documents and the question}.
This design inherently encourages the LLM to {internalize} the factual details in the documents into its parameters during training.
Although the generated question-answer pairs do not directly cover all the facts within the document, repeated training on the documents' tokens allows the model to reinforce its understanding of the textual content. 
Consequently, once the training is complete, the parametric representation \(\Delta \theta\) serves as a lightweight document-specific knowledge representation that can be directly added to the original model \(\mathcal{L}\) at inference time.

Notably, this entire process can be conducted offline, as each document (or batch of documents) is processed to produce its respective low-rank representation \(\Delta \theta\). 
At inference time, we only need to load the LoRA parameters corresponding to the specific document(s) rather than appending the document directly into the LLM’s context. 
The computational cost of loading these parametric representations constitutes only a minimal portion, approximately \textbf{$1\%$} of the computation required to decode a single token.

\subsubsection{Discussion on LoRA Initialization} \label{sec:lora-init-discussion}

In our proposed training framework, the LoRA parameters for each document $d_i$ are initialized randomly without any warm-up stage. This choice is deliberate and aligns with our goal of developing a general-purpose parametric knowledge injection method rather than one tailored to specific downstream tasks. 
If not explicitly mentioned otherwise, we initialize LoRA with random values.
However, randomized LoRA initialization is not necessarily the most effective way to train parametric document representations.
For example, we could pre-train the random LoRA with a couple of few-shot examples following the same method described with Eq. (\ref{eq:lm_training}) and save the LoRA weight $W_{warm-up}$ to initialize the training of each document's LoRA (i.e., the document's parametric representation).
Our experiment (Section \S ~\ref{ablation:init}) demonstrates that this warm-up process can significantly improve the performance compared to random initialization on RAG tasks, indicating that a \emph{task-aware} initialization can further enhance the effectiveness of parametric knowledge injection for specific downstream tasks.
Yet, we use random initialization for LoRA if not mentioned explicitly for simplicity and broad applicability across various tasks.

\subsection{Online Inference}
\label{sec:inference}

In the previous stage (\S\ref{sec:additional-param-training}), we generated a set of document-specific low-rank parameters for each document in the corpus \(K\). 
This section describes how these parameters are utilized for RAG pipelines. 
Given a user query \(q\), our proposed Parametric RAG pipeline proceeds through three steps: 
\textbf{Retrieve}, \textbf{Update}, and \textbf{Generate}.
The following section details each step and illustrates the underlying mathematical operations.

\subsubsection{Retrieve}
We first use a retriever \(R\) to calculate a relevance score \(S_{d_i}\) for each document \(d_i\in K\) to the query \(q\). 
We then select the top-\(k\) documents with the highest relevance scores, denoted as \(\{d_{1}, d_{2}, \dots, d_{k}\}\subseteq K\) as the relevant external knowledge. 
Each retrieved document \(d_{i}\) has a corresponding parametric representation, i.e., a pair of low-rank matrices \(\bigl(A_{i}, B_{i}\bigr)\), previously obtained by the procedure described in \S\ref{sec:additional-param-training}. 

\subsubsection{Update}
After retrieval, we merge the low-rank matrices from the top-\(k\) retrieved documents to form a single plug-in module for the LLM. 
Following the setting of LoRA~\cite{hulora} convention, we use a scalar scaling factor \(\alpha\) to modulate the final update. 
The {merged} weight update, \(\Delta W_{\text{merge}}\), is computed by summing over all retrieved documents:
\begin{align}
\label{eq:merged_update}
    \Delta W_{\text{merge}}
    &=\; \alpha \cdot
    \sum_{j=1}^{k} A_{j}\,B_{j}^\top
    .
\end{align}
Intuitively, \(\Delta W_{\text{merge}}\) combines the knowledge from multiple relevant documents into a single low-rank update that can be applied to the LLM’s base parameters.
Once we obtain \(\Delta W_{\text{merge}}\), we update the original feed-forward weight \(W\) by:
    $W' = W + \Delta W_{\text{merge}},$
thus yielding the final set of parameters for that layer at inference time. 
Conceptually, \(W'\) encodes the base model’s knowledge plus the aggregated knowledge from the top-\(k\) retrieved documents. 

\subsubsection{Generate}
After updating all feed-forward layers in the Transformer with \(\Delta W_{\text{merge}}\), we obtain a temporary model $\mathcal{L}'({\theta}')$, where \(\theta'\) represents the updated model parameters, which are obtained by incorporating the merged low-rank parameters for all retrieved documents.
We can then directly use \(\mathcal{L}'\) to generate the final response to the query \(q\) using a standard left-to-right decoding process.

\subsection{Discussion on Time/Space Efficiency}

\subsubsection{Computation Cost}
The computation cost of our method can be divided into offline preprocessing cost and online inference cost.
The offline cost primarily arises from the \textit{Parametric Document Encoding} (\S\ref{sec:additional-param-training}). 
Let $|d|$ be the average number of tokens in a document $d$, and \(h\) be the hidden dimension size of the LLM. 
The computational complexity of a typical decoder-only LLM is $\mathcal{O}(|d|^2h+|d|h^2)$, where the attention layers complexity is $\mathcal{O}(|d|^2h)$ plus the FFN layers $\mathcal{O}(|d|h^2)$.
Theoretically, our method only introduces a constant coefficient change on the number of tokens processed, thus the overall time complexity remains $\mathcal{O}(|d|^2h+|d|h^2)$.
Based on our implementation settings detailed in \S ~\ref{sec:implementation}, the Data Augmentation process takes the original document \( d \) as input and subsequently generates approximately \( 2|d| \) new tokens. 
This process requires computational costs equivalent to a forward pass over \( 3|d| \) tokens, including the decoding of \( 1|d| \) tokens and the inference of \( 2|d| \) tokens.
Training LoRA parameters on these augmented tokens requires a forward pass over $3|d|$ tokens and a backward pass equivalent to processing $6|d|$ tokens (typically about twice the forward-pass cost), resulting in an overall computational cost equivalent to processing $9|d|$ tokens. 
Adding the $3|d|$ tokens from the Document Augmentation phase, the total computational cost is approximately the cost of decoding $12|d|$ tokens in the LLM.


The online inference cost mainly depends on the number of input and output tokens. 
For simplicity, we focus on input tokens and ignore the variance in output tokens since they vary significantly from tasks and LLMs. 
Let $|q|$ represent the length of input prompt/question $q$, and $t$ be the number of retrieved documents.
Since the time needed to load the LoRA parameters for $t$ documents is neglectable, the inference time complexity of our method is $\mathcal{O}(|q|^2h+|q|h^2)$.
In contrast, the time complexity of in-context RAG methods is $\mathcal{O}\big((t|d|+|q|)^2h+(t|d|+|q|)h^2\big)$, which means that our method can save $\mathcal{O}\big(t^2|d|^2h+t|d||q|h+t|d|h^2\big)$ time for online inference.
Empirically, suppose that the lengths of $q$ and $d$ are approximately the same and significantly smaller than the hidden dimension of the LLM (e.g., about 4096 for LLaMA-8B), and we retrieve $t=6$ documents for each $q$, then our method can roughly save $6|d|$ tokens in inference.  
Compared to its offline cost, this means that Parametric RAG is more cost-friendly than in-context RAG when the number of queries is more than twice that of documents in the life cycle of the service.

In summary, while the offline preprocessing step in Parametric RAG introduces additional computational overhead compared to traditional RAG, our analysis demonstrates that, when the system handles a large number of queries, Parametric RAG can provide a more carbon-efficient solution for large-scale RAG systems.

\subsubsection{Storage Overhead}
In Parametric RAG, storage overhead comes from the Parametric Representation of each document, which consists of low-rank matrices from the FFN layer. Let \( r \) be the LoRA rank, $n$ be the number of Transformer layers, $h$ be the hidden size, and \( l \) the intermediate size of FFN, then the number of parameters in the Parametric Representation of a document is \( 2nr(h+l) \).
For example, with the LLaMA3-8B model (32 layers, hidden size 4096, intermediate size 14336), we need to store approximately 2.36M extra parameters (with \( r = 2 \) as used in our experiments). Stored at 16-bit precision, this requires around 4.72MB per document.

While the storage requirements for Parametric RAG may seem substantial compared to the raw documents, there are multiple methods to reduce its cost in practice.
For example, previous studies find that the access of information in real user traffic follows a long-tail distribution~\cite{silver1998}.
Taking Google as an example, about 96.55\% of Web pages receive zero traffic, and only 1.94\% get one to ten visits per month~\cite{soulo2023traffic}.
Therefore, creating parametric representations for a tiny set of head documents can serve the majority of user requests, which significantly reduces the storage cost of Parametric RAG.
Also, as shown in our experiments, Parametric RAG can be used with in-context RAG together for downstream tasks.
Thus, it can serve as a natural boost to existing RAG methods without breaking their system pipelines.

\section{Experimental Setup}

In this section, we detail the experimental framework used to evaluate Parametric RAG. We begin with the introduction of our selected benchmark datasets(\S\ref{sec:benchmarks}). 
Next, we introduce our selected baseline methods (\S\ref{sec:baselines}) and implementation. 
Finally, we provide implementation details regarding our parameterization method, retrieval strategy, and inference settings (\S\ref{sec:implementation}).

\subsection{Benchmarks and Metrics}
\label{sec:benchmarks}
We evaluate our approach on diverse benchmark datasets, each designed to assess different reasoning capabilities, such as multi-hop reasoning and commonsense inference. Specifically, we select the following datasets:

\begin{itemize}[leftmargin=*]

\item\textbf{2WikiMultihopQA (2WQA)}~\cite{ho2020constructing} is a dataset designed to test the model’s ability to perform multi-hop reasoning by integrating information across multiple Wikipedia passages.

\item\textbf{HotpotQA (HQA)}~\cite{yang2018hotpotqa} also focuses on evaluating multi-hop reasoning skills, requiring models to combine information from different contexts to address a single query.

\item\textbf{PopQA (PQA)}~\cite{mallen-etal-2023-trust} assesses factual question answering, challenging the model’s ability to recall accurate knowledge and resolve ambiguity in entity representation.

\item\textbf{ComplexWebQuestions (CWQ)}~\cite{talmor-berant-2018-web} involves answering multi-step, web-based questions, further testing the model’s capacity to retrieve and reason over large-scale web content.

\end{itemize}
\noindent For evaluation metrics, we use the F1 score to evaluate performance on question-answering tasks, as it captures the balance between precision and recall by accounting for partially correct answers. Both \textbf{2WQA} and \textbf{HQA} categorize questions based on reasoning types, with 2WQA divided into four categories and HQA into two. To comprehensively compare the performance of P-RAG and other RAG baselines across different reasoning tasks, our main experimental table (Table~\ref{table:main}) presents the performance for each sub-task separately, using the first 300 questions from each sub-dataset. Table~\ref{table:main} also presents the overall performance of each RAG baseline on the two datasets in the ``Total'' column. Since the original datasets contain uneven distributions of question types, the ``Total'' column is not a simple average of the sub-dataset performances.

\begin{table*}[t]
\caption{The overall experiment results of Parametric RAG and other baselines across four tasks.
All metrics reported are F1 scores.
Bold numbers indicate the best performance of all baselines, and the second-best results are underlined. 
“*” and $\dagger$ denote significantly worse performance than the bolded method and our proposed P-RAG with $p < 0.05$ level, respectively.}
\label{table:main}
\centering
\setlength\tabcolsep{2.8pt}
\begin{tabular}{|cl|c|c|c|c|c|c|c|c|c|c|}
\toprule

\multirow{2}{*}{\textbf{}} & \multirow{2}{*}{\makecell[c]{\textbf{}}} & \multicolumn{5}{c}{\textbf{2WikiMultihopQA}}

  & \multicolumn{3}{|c|}{\textbf{HotpotQA}} 
  & \multirow{2}{*}{\textbf{PopQA}} 
  & \multirow{2}{*}{\textbf{CWQ}} \\
\cmidrule(lr){3-7}\cmidrule(lr){8-10}
&  & \textbf{Compare} & \textbf{Bridge} & \textbf{Inf.} & \textbf{Compose} & \textbf{Total}
  & \textbf{Bridge}  & \textbf{Compare} & \textbf{Total}
  & & \\
\toprule

\multirow{6}{*}{\textbf{LLaMA-1B}}

& \textbf{Standard RAG} 
& \multicolumn{1}{l|}{~0.4298\textsuperscript{\tiny $\dagger$}*}& \multicolumn{1}{l|}{~0.3032\textsuperscript{\tiny $\dagger$}*}& \multicolumn{1}{l|}{~0.2263}& \multicolumn{1}{l|}{~0.1064} & \multicolumn{1}{l|}{~0.2520*}& \multicolumn{1}{l|}{~\underline{0.2110}}& \multicolumn{1}{l|}{~0.4083} & \multicolumn{1}{l|}{~\underline{0.2671}}& \multicolumn{1}{l|}{~0.1839*}& \multicolumn{1}{l|}{~0.3726}\\

& \textbf{DA-RAG} 
& \multicolumn{1}{l|}{~0.3594\textsuperscript{\tiny $\dagger$}*}& \multicolumn{1}{l|}{~0.2587\textsuperscript{\tiny $\dagger$}*}& \multicolumn{1}{l|}{~\underline{0.2266}}& \multicolumn{1}{l|}{~0.0869\textsuperscript{\tiny $\dagger$}*}& \multicolumn{1}{l|}{~0.2531*}& \multicolumn{1}{l|}{~0.1716*}& \multicolumn{1}{l|}{~0.3713\textsuperscript{\tiny $\dagger$}*}& \multicolumn{1}{l|}{~0.2221}& \multicolumn{1}{l|}{~0.2012*}& \multicolumn{1}{l|}{~0.3691}\\

& \textbf{FLARE} 
& \multicolumn{1}{l|}{~0.4013\textsuperscript{\tiny $\dagger$}*}& \multicolumn{1}{l|}{~0.2589\textsuperscript{\tiny $\dagger$}*}& \multicolumn{1}{l|}{~0.1960}& \multicolumn{1}{l|}{~0.0823\textsuperscript{\tiny $\dagger$}*}& \multicolumn{1}{l|}{~0.2234*}& \multicolumn{1}{l|}{~0.1630*}& \multicolumn{1}{l|}{~0.3784\textsuperscript{\tiny $\dagger$}*}& \multicolumn{1}{l|}{~0.1785*}& \multicolumn{1}{l|}{~0.1301\textsuperscript{\tiny $\dagger$}*}& \multicolumn{1}{l|}{~0.3173*}\\

& \textbf{DRAGIN}
& \multicolumn{1}{l|}{~0.4556}& \multicolumn{1}{l|}{~0.3357*}& \multicolumn{1}{l|}{~0.1919}& \multicolumn{1}{l|}{~0.0901}& \multicolumn{1}{l|}{~0.2692*}& \multicolumn{1}{l|}{~0.1431*}& \multicolumn{1}{l|}{~0.4015}& \multicolumn{1}{l|}{~0.1830*}& \multicolumn{1}{l|}{~0.1056\textsuperscript{\tiny $\dagger$}*}& \multicolumn{1}{l|}{~\underline{0.3900}}\\

& \textbf{P-RAG  (Ours)}
& \multicolumn{1}{l|}{~\underline{0.4920}}& \multicolumn{1}{l|}{~\underline{0.3994}}& \multicolumn{1}{l|}{~0.2185}& \multicolumn{1}{l|}{~\underline{0.1334}}& \multicolumn{1}{l|}{~\underline{0.2764}}& \multicolumn{1}{l|}{~0.1602*}& \multicolumn{1}{l|}{~\textbf{0.4493}}& \multicolumn{1}{l|}{~0.1999*}& \multicolumn{1}{l|}{~\underline{0.2205}*}& \multicolumn{1}{l|}{~0.3482*}\\

& \textbf{Combine Both}
& \multicolumn{1}{l|}{~\textbf{0.5046}}& \multicolumn{1}{l|}{~\textbf{0.4595}}& \multicolumn{1}{l|}{~\textbf{0.2399}}& \multicolumn{1}{l|}{~\textbf{0.1357}}& \multicolumn{1}{l|}{~\textbf{0.3237}}& \multicolumn{1}{l|}{~\textbf{0.2282}}& \multicolumn{1}{l|}{~\underline{0.4217}}& \multicolumn{1}{l|}{~\textbf{0.2689}}
& \multicolumn{1}{l|}{~\textbf{0.2961}}& \multicolumn{1}{l|}{~\textbf{0.4101}}\\

\toprule

\multirow{6}{*}{\textbf{Qwen-1.5B}}


& \textbf{Standard RAG}
& \multicolumn{1}{l|}{~0.3875\textsuperscript{\tiny $\dagger$}*}& \multicolumn{1}{l|}{~0.3884\textsuperscript{\tiny $\dagger$}*}& \multicolumn{1}{l|}{~0.1187\textsuperscript{\tiny $\dagger$}*}& \multicolumn{1}{l|}{~0.0568\textsuperscript{\tiny $\dagger$}*}& \multicolumn{1}{l|}{~0.2431\textsuperscript{\tiny $\dagger$}*}& \multicolumn{1}{l|}{~0.1619*}& \multicolumn{1}{l|}{~0.3713\textsuperscript{\tiny $\dagger$}*}& \multicolumn{1}{l|}{~0.2073*}& \multicolumn{1}{l|}{~0.0999\textsuperscript{\tiny $\dagger$}*}& \multicolumn{1}{l|}{~0.2823*}\\

& \textbf{DA-RAG} 
& \multicolumn{1}{l|}{~0.3418\textsuperscript{\tiny $\dagger$}*}& \multicolumn{1}{l|}{~0.4015}& \multicolumn{1}{l|}{~0.1269\textsuperscript{\tiny $\dagger$}*}& \multicolumn{1}{l|}{~0.0514\textsuperscript{\tiny $\dagger$}*}& \multicolumn{1}{l|}{~0.2156\textsuperscript{\tiny $\dagger$}*}& \multicolumn{1}{l|}{~0.1182\textsuperscript{\tiny $\dagger$}*}& \multicolumn{1}{l|}{~0.3041\textsuperscript{\tiny $\dagger$}*}& \multicolumn{1}{l|}{~0.1683*}& \multicolumn{1}{l|}{~0.1197\textsuperscript{\tiny $\dagger$}*}& \multicolumn{1}{l|}{~0.2718\textsuperscript{\tiny $\dagger$}*}\\

& \textbf{FLARE} 
& \multicolumn{1}{l|}{~0.1896\textsuperscript{\tiny $\dagger$}*}& \multicolumn{1}{l|}{~0.1282\textsuperscript{\tiny $\dagger$}*}& \multicolumn{1}{l|}{~0.0852\textsuperscript{\tiny $\dagger$}*}& \multicolumn{1}{l|}{~0.0437\textsuperscript{\tiny $\dagger$}*}& \multicolumn{1}{l|}{~0.1004\textsuperscript{\tiny $\dagger$}*}& \multicolumn{1}{l|}{~0.0750\textsuperscript{\tiny $\dagger$}*}& \multicolumn{1}{l|}{~0.1229\textsuperscript{\tiny $\dagger$}*}& \multicolumn{1}{l|}{~0.0698\textsuperscript{\tiny $\dagger$}*}& \multicolumn{1}{l|}{~0.0641\textsuperscript{\tiny $\dagger$}*}& \multicolumn{1}{l|}{~0.1647\textsuperscript{\tiny $\dagger$}*}\\

& \textbf{DRAGIN}
& \multicolumn{1}{l|}{~0.2771\textsuperscript{\tiny $\dagger$}*}& \multicolumn{1}{l|}{~0.1826\textsuperscript{\tiny $\dagger$}*}& \multicolumn{1}{l|}{~0.1025\textsuperscript{\tiny $\dagger$}*}& \multicolumn{1}{l|}{~0.0680\textsuperscript{\tiny $\dagger$}*}& \multicolumn{1}{l|}{~0.1538\textsuperscript{\tiny $\dagger$}*}& \multicolumn{1}{l|}{~0.0801\textsuperscript{\tiny $\dagger$}*}& \multicolumn{1}{l|}{~0.1851\textsuperscript{\tiny $\dagger$}*}& \multicolumn{1}{l|}{~0.0973\textsuperscript{\tiny $\dagger$}*}& \multicolumn{1}{l|}{~0.0548\textsuperscript{\tiny $\dagger$}*}& \multicolumn{1}{l|}{~0.1788\textsuperscript{\tiny $\dagger$}*}\\

& \textbf{P-RAG (Ours)}
& \multicolumn{1}{l|}{~\textbf{0.4529}}& \multicolumn{1}{l|}{~\textbf{0.4494}}& \multicolumn{1}{l|}{~\textbf{0.2072}}& \multicolumn{1}{l|}{~\textbf{0.1372}}& \multicolumn{1}{l|}{~\textbf{0.3025}}& \multicolumn{1}{l|}{~\underline{0.1720}}& \multicolumn{1}{l|}{~\underline{0.4623}}& \multicolumn{1}{l|}{~\underline{0.2165*}}& \multicolumn{1}{l|}{~\underline{0.1885}}& \multicolumn{1}{l|}{~\underline{0.3280}}\\

& \textbf{Combine Both}
& \multicolumn{1}{l|}{~\underline{0.4053}}& \multicolumn{1}{l|}{~\underline{0.4420}}& \multicolumn{1}{l|}{~\underline{0.1705}}& \multicolumn{1}{l|}{~\underline{0.1154}}& \multicolumn{1}{l|}{~\underline{0.2627}}& \multicolumn{1}{l|}{~\textbf{0.2383}}& \multicolumn{1}{l|}{~\textbf{0.5037}}& \multicolumn{1}{l|}{~\textbf{0.2942}}& \multicolumn{1}{l|}{~\textbf{0.2261}}& \multicolumn{1}{l|}{~\textbf{0.3495}}\\

\toprule

\multirow{6}{*}{\textbf{LLaMA-8B}}

& \textbf{Standard RAG}
& \multicolumn{1}{l|}{~0.5843\textsuperscript{\tiny $\dagger$}*}& \multicolumn{1}{l|}{~0.4794\textsuperscript{\tiny $\dagger$}*}& \multicolumn{1}{l|}{~0.1833\textsuperscript{\tiny $\dagger$}*}& \multicolumn{1}{l|}{~0.0991\textsuperscript{\tiny $\dagger$}*}& \multicolumn{1}{l|}{~0.3372\textsuperscript{\tiny $\dagger$}*}& \multicolumn{1}{l|}{~0.1823\textsuperscript{\tiny $\dagger$}*}& \multicolumn{1}{l|}{~0.3493\textsuperscript{\tiny $\dagger$}*}& \multicolumn{1}{l|}{~0.2277\textsuperscript{\tiny $\dagger$}*}& \multicolumn{1}{l|}{~0.1613\textsuperscript{\tiny $\dagger$}*}& \multicolumn{1}{l|}{~0.3545\textsuperscript{\tiny $\dagger$}*}\\

& \textbf{DA-RAG} 
& \multicolumn{1}{l|}{~0.4921\textsuperscript{\tiny $\dagger$}*}& \multicolumn{1}{l|}{~0.3344\textsuperscript{\tiny $\dagger$}*}& \multicolumn{1}{l|}{~0.1523\textsuperscript{\tiny $\dagger$}*}& \multicolumn{1}{l|}{~0.0670\textsuperscript{\tiny $\dagger$}*}& \multicolumn{1}{l|}{~0.2396\textsuperscript{\tiny $\dagger$}*}& \multicolumn{1}{l|}{~0.1587\textsuperscript{\tiny $\dagger$}*}& \multicolumn{1}{l|}{~0.2860\textsuperscript{\tiny $\dagger$}*}& \multicolumn{1}{l|}{~0.1996\textsuperscript{\tiny $\dagger$}*}& \multicolumn{1}{l|}{~0.2255*}& \multicolumn{1}{l|}{~0.3481\textsuperscript{\tiny $\dagger$}*}\\

& \textbf{FLARE} 
& \multicolumn{1}{l|}{~0.4293\textsuperscript{\tiny $\dagger$}*}& \multicolumn{1}{l|}{~0.3769\textsuperscript{\tiny $\dagger$}*}& \multicolumn{1}{l|}{~\underline{0.3086}}& \multicolumn{1}{l|}{~0.1627*}& \multicolumn{1}{l|}{~0.3492*}& \multicolumn{1}{l|}{~0.2493\textsuperscript{\tiny $\dagger$}*}& \multicolumn{1}{l|}{~0.4324\textsuperscript{\tiny $\dagger$}*}& \multicolumn{1}{l|}{~0.2771\textsuperscript{\tiny $\dagger$}*}& \multicolumn{1}{l|}{~0.2393*}& \multicolumn{1}{l|}{~0.3084\textsuperscript{\tiny $\dagger$}*}\\

& \textbf{DRAGIN}
& \multicolumn{1}{l|}{~0.5185\textsuperscript{\tiny $\dagger$}*}& \multicolumn{1}{l|}{~0.4480\textsuperscript{\tiny $\dagger$}*}& \multicolumn{1}{l|}{~0.2664}& \multicolumn{1}{l|}{~0.1833}& \multicolumn{1}{l|}{~0.3544*}& \multicolumn{1}{l|}{~0.2618*}& \multicolumn{1}{l|}{~0.6116*}& \multicolumn{1}{l|}{~0.2924*}& \multicolumn{1}{l|}{~0.1772\textsuperscript{\tiny $\dagger$}*}& \multicolumn{1}{l|}{~0.3101\textsuperscript{\tiny $\dagger$}*}\\

& \textbf{P-RAG  (Ours)}
& \multicolumn{1}{l|}{~\underline{0.6353}}& \multicolumn{1}{l|}{~\underline{0.5437}}& \multicolumn{1}{l|}{~0.2471*}& \multicolumn{1}{l|}{~\underline{0.1992}}& \multicolumn{1}{l|}{~\underline{0.3932}}& \multicolumn{1}{l|}{~\underline{0.3115}*}& \multicolumn{1}{l|}{~\underline{0.6557}}& \multicolumn{1}{l|}{~\underline{0.3563}*}& \multicolumn{1}{l|}{~\underline{0.2413}*}& \multicolumn{1}{l|}{~\underline{0.4541}}\\

& \textbf{Combine Both}
& \multicolumn{1}{l|}{~\textbf{0.6432}}& \multicolumn{1}{l|}{~\textbf{0.5556}}& \multicolumn{1}{l|}{~\textbf{0.3160}}& \multicolumn{1}{l|}{~\textbf{0.2339}}& \multicolumn{1}{l|}{~\textbf{0.4258}}& \multicolumn{1}{l|}{~\textbf{0.4025}}& \multicolumn{1}{l|}{~\textbf{0.6918}}& \multicolumn{1}{l|}{~\textbf{0.4559}}& \multicolumn{1}{l|}{~\textbf{0.3059}}& \multicolumn{1}{l|}{~\textbf{0.4728}}\\

\bottomrule
\end{tabular}

\end{table*}

\subsection{Baselines}
\label{sec:baselines}

We choose the following RAG baselines for comparison: 

\begin{itemize}[leftmargin=*]

\item \textbf{Standard RAG}. 
This RAG method directly appends the top retrieved documents to the LLM’s input prompt. The prompt explicitly instructs the LLM to refer to the provided documents when answering the question and also includes instructions on the output format expected from the model.

\item \textbf{DA-RAG} incorporates the augmented documents and QA pairs using the Data Augmentation method introduced in \S ~\ref{sec:DA}. 
This baseline aims to demonstrate that the performance improvement observed in Parametric RAG does not stem from the data augmentation phase but from the in-parameter knowledge injection.

\item \textbf{FLARE}~\cite{jiang2023active} is a multi-round retrieval augmentation method that triggers retrieval each time it encounters an uncertain token. When the retrieval module is triggered, the last generated sentence without the uncertain tokens is defined as the query.

\item \textbf{DRAGIN}~\cite{su2024dragin} is a multi-round retrieval augmentation method. It triggers retrieval when an uncertain token has semantic meaning and also has a strong influence on the following tokens. When the retrieval module is triggered, it formulates the query based on the model’s internal state and preceding context.

\item \textbf{P-RAG} directly injects relevant documents into the LLM’s parameters through document parameterization, enabling efficient RAG without increasing the input context length.

\item \textbf{Combine Both}. This baseline combines the in-context RAG method with P-RAG, leveraging both in-context and parametric knowledge injection. 
This baseline aims to evaluate whether the fusion of these approaches leads to better performance.

\end{itemize}

For P-RAG and all the baselines, we use the same retriever and select the top $3$ retrieved documents as relevant.
To ensure a fair comparison, we ensured that P-RAG and all the baselines share the same prompt template\footnote{All the prompt templates used in this paper are available in our GitHub repository: https://github.com/oneal2000/PRAG/blob/main/all\_prompt.md} under the same dataset.

\subsection{Implementation Details}
\label{sec:implementation}

In this subsection, we introduce the specific implementation of our experiments:

\textbf{Base Models}
We implement Parametric RAG using open-source pre-trained LLMs. 
To ensure the broad effectiveness of P-RAG across different models, we selected LLMs of varying scales and from different series, including Qwen2.5-1.5B-Instruct~\citep{yang2024qwen2}, LLaMA-3.2-1B-Instruct~\citep{Llama-3.2-1B-Instruct}, and Llama-3-8B-Instruct~\cite{Llama-3-8B-Instruct}. All experiments were conducted using PyTorch on NVIDIA A100 GPUs with 40GB of memory.

\textbf{Preprocessing and Parameterization.} 
Consistent with prior works~\cite{karpukhin2020dense,jiang2023active,su2024dragin,su2024mitigating}, we utilize Wikipedia dumps as our external knowledge corpus, specifically adopting the dataset\footnote{https://github.com/facebookresearch/DPR/tree/main} proposed by DPR~\cite{karpukhin2020dense}.
For document augmentation, each document is rewritten once, and three QA pairs are generated based on the document\footnote{The detailed prompt template for document augmentation is publicly available on our official GitHub repository.} (using the downstream LLM, if not mentioned explicitly).
In the LoRA fine-tuning process, the learning rate was set to $3 \times 10^{-4}$, and the training epoch was set to $1$. The LoRA modules were exclusively integrated into the feed-forward network (FFN) matrices, excluding the query, key, and value ($QKV$) matrices. 
The scaling factor $\alpha$ was configured to 32, LoRA rank $r$ was set to 2, and no dropout was applied during training to ensure stability and full utilization of the parameter updates.
The LoRA weight is randomly initialized following the setting of the original LoRA paper~\cite{hulora}.

\textbf{Retrieval Module.} 
Recent studies on retrieval-augmented generation (RAG)~\cite{ram2023context} reveal that BM25 performs on par with, or even outperforms, state-of-the-art dense models in some scenarios. 
Given its strong performance, simplicity, and low computational cost, we adopt BM25 as the retriever for our approach.
We use Elasticsearch as the backend for implementing BM25, with detailed configuration settings and instructions available on our official GitHub repository.

\textbf{Generation Configuration.} 
All experiments are conducted using the publicly released Hugging Face implementations of LLaMA and Qwen. 
We adopt the default hyperparameters and chat template provided in the official Huggingface repository, with the only modification being the use of greedy decoding to ensure the reproducibility of our reported results.

\section{Experiments}

\subsection{Main Experiment}

In this section, we present the main experimental result and an in-depth analysis of our proposed Parametric RAG compared with other RAG baselines and a combined setting that leverages both parametric and in-context knowledge injection. 
The experimental results are presented in Table~\ref{table:main}, and we provide the following analysis:
\textbf{(1) Overall Analysis.} 
P-RAG outperforms existing RAG frameworks in most of the benchmarks and LLMs evaluated. This trend is especially obvious for Qwen-1.5B and LLaMA-8B. The improvements suggest that the incorporation of knowledge into model parameters can enhance the overall performance of the RAG pipeline, enabling the model to recall and reason over the injected knowledge more effectively. Furthermore, since these gains are observed in models from different series and parameter sizes, the results underscore the robustness and broad applicability of P-RAG.
\textbf{(2) Comparison with DA-RAG.}
DA-RAG incorporates all the content generated during the Document Augmentation phase into the context, whereas our proposed P-RAG consistently outperforms DA-RAG across all settings. This result demonstrates that the performance improvement observed in Parametric RAG does not stem from the document augmentation phase, but from the in-parameter knowledge injection paradigm.
\textbf{(3) Impact of Model Scale on P-RAG.} 
The performance gap between P-RAG and other RAG baselines is noticeably more significant when moving from the LLaMA-1B model to LLaMA-8B. This discrepancy indicates that parametric injection becomes even more beneficial in larger-scale models because larger models can better leverage internalized document knowledge. 
\textbf{(4) Combine In-context RAG and P-RAG.} 
The combined use of parametric and in-context RAG methods (Combine Both) yields the highest overall performance across various datasets and base LLMs. 
This result highlights that in-parameter knowledge injection is not in conflict with traditional RAG methods based on in-context knowledge injection.
Consequently, our proposed document parameterization approaches can be seamlessly integrated for downstream tasks, allowing Parametric RAG to enhance existing RAG systems without disrupting their pipelines.
\textbf{(5) Other Findings.}
Both DRAGIN and FLARE underperform significantly when applied to Qwen-2.5-1.5B. Our analysis suggests that Qwen-2.5-1.5B tends to produce highly confident answers regardless of uncertainty. Since these dynamic RAG frameworks rely on confidence to trigger retrieval, they rarely activate on Qwen-2.5-1.5B. 
This highlights a key limitation of uncertainty-based triggers and underscores the need for more robust mechanisms in dynamic RAG frameworks.

\subsection{Impact of LoRA Weight Initialization}
\label{ablation:init}

\begin{table}
\caption{Ablation study on the impact of LoRA weight initialization strategies for P-RAG. All metrics reported are F1 scores. 
“P-RAG Rand.” and “P-RAG Warm.” indicate randomly initialized LoRA weights and warm-up LoRA initialization, respectively. The best results are in bold.
}

\label{table:init}
\centering
\setlength\tabcolsep{2.5pt}
\begin{tabular}{clcccc}
\toprule
                                    &                       & \textbf{2WQA}   & \textbf{HQA} & \textbf{PQA}  & \textbf{CWQ}     \\
                                    \toprule
\multirow{2}{*}{\textbf{LLaMA-1B}}  
                                    & \textbf{P-RAG Rand.}  & 0.2764          & 0.1999            & \textbf{0.2205} & 0.3482           \\
                                    & \textbf{P-RAG Warm.} & \textbf{0.3546} & \textbf{0.2456}            & 0.2035          & \textbf{0.4263}  \\
                                    \toprule
\multirow{2}{*}{\textbf{Qwen-1.5B}} 
                                    & \textbf{P-RAG Rand.}  & 0.3025          & 0.2165            & 0.1885          & 0.3280           \\
                                    & \textbf{P-RAG Warm.} & \textbf{0.3542} & \textbf{0.2718}   & \textbf{0.2418} & \textbf{0.5018}  \\
                                    \toprule
\multirow{2}{*}{\textbf{LLaMA-8B}}  
                                    & \textbf{P-RAG Rand.}  & 0.3932          & 0.3563            & 0.2413          & 0.4541           \\
                                    & \textbf{P-RAG Warm.} & \textbf{0.4201} & \textbf{0.4499}   & \textbf{0.2952} & \textbf{0.5591}  \\
                                    \toprule
\end{tabular}

\end{table}

To investigate the impact of LoRA weight initialization strategies on our proposed Parametric RAG framework, we conducted an ablation study using two initialization strategies:


\noindent (1) \textbf{Random Initialization} (P-RAG Rand.): 
The LoRA weights are initialized randomly without any pretraining or warm-up, which represents the default setting for our proposed Parametric RAG approach. 
{All the Parametric RAG experimental results reported in other sections in this paper are based on this setting.}

\noindent (2) \textbf{Warm-Up Initialization} (P-RAG Warm.):  LoRA weights are pre-trained using a set of 600 sampled question-answer (QA) pairs. 
These QA pairs are selected from the training sets of our chosen benchmarks and are distinct from the test questions to ensure no data leakage. 
The pre-training process involves training the LoRA parameters using the standard next-token prediction method on the concatenated tokens of the QA pairs. 
The specific implementation follows Section \ref{sec:implementation}, and the pre-trained LoRA parameters are saved as initialization for the Document Parameterization phase\footnote{All the training code and data are publicly available at our anonymous GitHub repository: https://github.com/oneal2000/PRAG}.


The experimental results in Table~\ref{table:init} clearly indicate that across different model series and scales, as well as diverse datasets, the warm-up initialization strategy (P-RAG Warm.) consistently outperforms random initialization (P-RAG Rand.). 
This demonstrates the effectiveness of task-aware pretraining in enhancing the Parametric RAG pipeline. 
Furthermore, the observed improvements across varying model sizes confirm the scalability and generality of this approach. 
The superior performance of the warm-up approach in downstream tasks can be attributed to two key factors. First, it effectively aligns the additional LoRA parameters with the base LLM before document parameterizing, ensuring a smoother integration of knowledge. 
Second, it facilitates the incorporation of task-relevant knowledge, including output formats and generation patterns, which are critical for enhancing the quality of response in certain tasks.
This finding suggests that in practical Parametric RAG applications where the downstream task is fixed, warming up the LoRA parameters for the task offers a promising approach to boost effectiveness.
{It is important to note that our main experiments (as well as all other experiments in this paper) were conducted using random initialization without any task-specific optimizations or dataset-specific tuning. This further highlights the strong generalization capability of our proposed Parametric RAG paradigm. }

These findings also highlight a broader insight: 
embedding few-shot examples either in the model's context or directly into its parameters leads to improved downstream task performance. 
Interestingly, our proposed parametric information representation method offers compatibility with few-shot in-context learning, enabling a combination of parametric and in-context knowledge augmentation.

\subsection{Impact of Document Augmentation}

\begin{figure}[t]
\centering
    \includegraphics[width=\columnwidth]{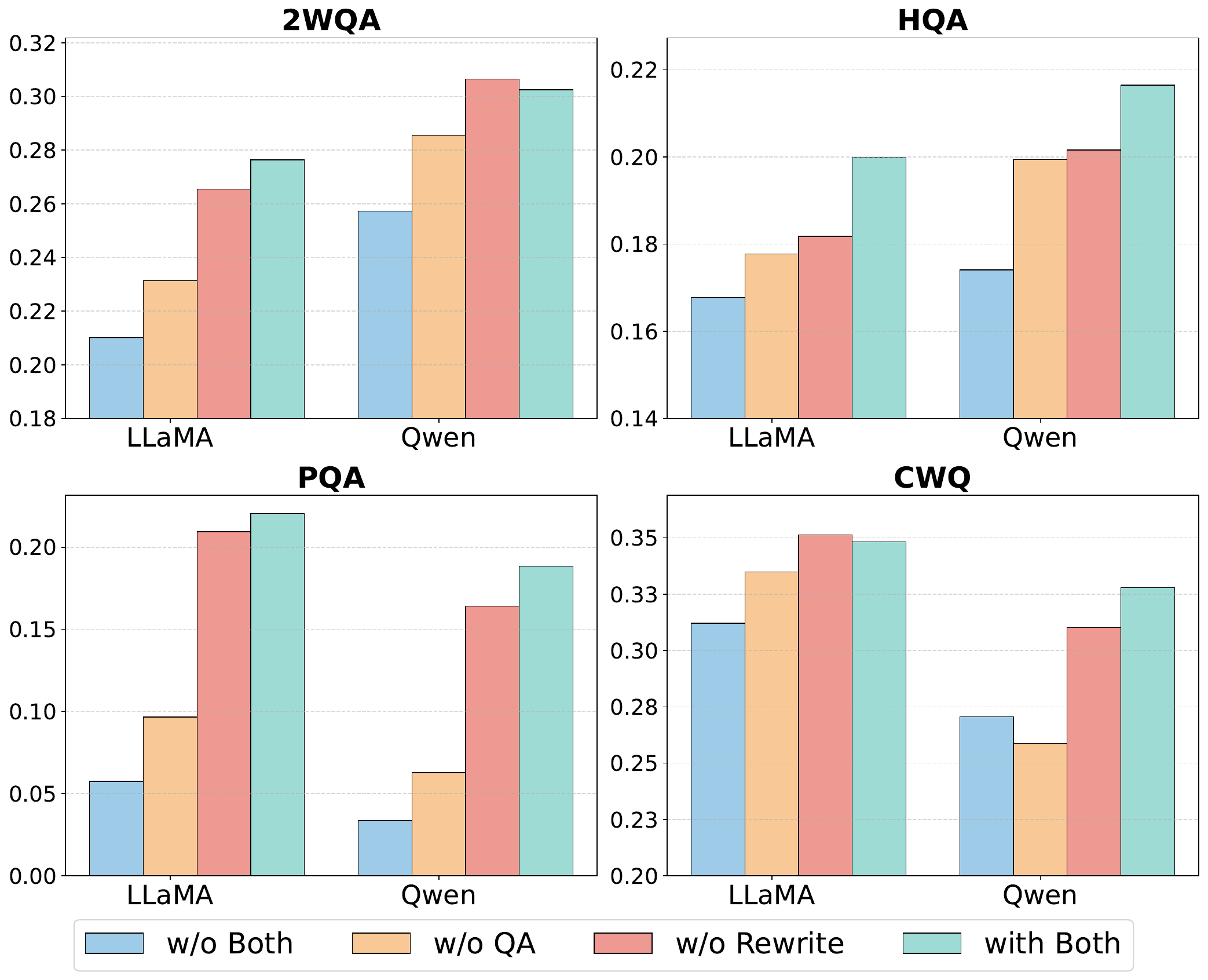}
    \caption{Ablation study on the impact of the document augmentation stage. LLaMA indicates LLaMA-3.2-1B, and Qwen indicates Qwen-2.5-1.5B. The metric used is the F1 Score.}
    \label{pic:exp}

\end{figure}

To investigate the individual contributions of the rewriting and question-answer (QA) generation steps in the document augmentation process, we conduct a series of ablation experiments by removing (1) both rewriting and QA, (2) QA alone, and (3) rewriting alone. 
The experimental results are shown in Figure~\ref{pic:exp}, and we have the following observations:
\textbf{(1)} When neither rewriting nor QA generation is employed, the performance consistently degrades significantly across all evaluated tasks and models. This reduction suggests that simply training the LLM on the selected document via the next token prediction task without any form of data augmentation leads to insufficient internalization of facts by the model. \textbf{(2)} Removing either QA or rewriting alone yields better results than removing both, indicating that each step offers distinct benefits. 
However, we notice that removing QA leads to a more significant performance decline than removing rewriting.
This observation suggests that QA pair generation is more crucial for pushing the model to recall and apply factual information while rewriting offers valuable diversity in phrasing and structure and benefits the overall performance. \textbf{(3)} Incorporating both rewriting and QA results in the strongest overall performance on most of the evaluated tasks and models. These findings reinforce that rewriting and QA generation play complementary roles.

In general, this ablation study indicates that both rewriting and QA generation significantly enhance the performance of document parameterizing. Their integration produces the best performance. Rewriting expands the coverage and diversity of context, while QA explicitly encourages the model to encode the knowledge of the selected document in a necessary way to apply the knowledge for downstream tasks. Therefore, it is advisable to incorporate both components of our document augmentation for effective internalization of knowledge.

\subsection{Impact of Data-augmentation Model}
\label{ablation:model}

\label{ablation:self}
\begin{table}
\caption{Ablation study comparing different document augmentation models. GenLM indicates the generator LLM and AugLM indicates the LLM for document augmentation. LLaMA indicates LLaMA-3.2-1B, and Qwen indicates Qwen-2.5-1.5B. The best results are in bold. The metric used in the table is F1 Score.}
\label{table:augLM}
\centering
\begin{tabular}{clcccc}
\toprule
\multirow{2}{*}{\textbf{GenLM }} & \multicolumn{1}{c}{\multirow{2}{*}{\textbf{AugLM }}} & \multicolumn{4}{c}{\textbf{Dataset }}                       \\
\cmidrule(lr){3-6}
                                   & \multicolumn{1}{c}{}                                   & \textbf{2WQA} & \textbf{HQA} & \textbf{PQA} & \textbf{CWQ}  \\
                                   \toprule
\multirow{3}{*}{\textbf{LLaMA-1B}}& \textbf{LLaMA-1B}& 0.2764& 0.1999& 0.2205& 0.3482
\\
                                   & \textbf{Qwen-1.5B}& 0.2753& 0.1980& 0.2340& 0.3495
\\
                                   & \textbf{LLaMA-8B}& 0.2748& 0.1935& 0.2207& 0.3498\\
                                   \toprule
\multirow{3}{*}{\textbf{Qwen-1.5B}}& \textbf{LLaMA-1B}& 0.2974& 0.2005& 0.1829& 
0.3183\\
                                   & \textbf{Qwen-1.5B}& 0.3025& 0.2165& 0.1885& 0.3280\\
                                   & \textbf{LLaMA-8B}& 0.2948& 0.2161& 0.2156& 0.3211\\
                                   \toprule
\end{tabular}
\vspace{-5mm}
\end{table}


To evaluate the impact of the choice of LLM in the Document Augmentation phase, we conducted an ablation study comparing different configurations of the model used for document rewriting and QA pair generation. In our default setting, we use the same LLM for both the document augmentation process and the downstream task. However, to explore whether the performance of our method is sensitive to this choice, we tested alternative configurations. 
Specifically, we tested different model sizes by using both smaller and larger LLMs for document augmentation and QA pair generation.
The experimental results are shown in Table~\ref{table:augLM}, indicating that our framework demonstrates an insensitivity to the choice of data augmentation model. 
Performance remains consistent across different setups, regardless of whether a smaller, larger, or the same model as the generator is used for data augmentation. Importantly, using a small model for augmentation yields comparable results to employing significantly larger models, indicating that the augmentation step does not require high-capacity models to be effective. Similarly, when the same model is used for both generation and augmentation, the outcomes are indistinguishable from those where separate models are employed.

\subsection{Runtime Analysis}

We present the inference time for the LLaMA3-8B model across various RAG baselines on 2WikiMultihopQA (2WQA) and ComplexWebQuestions (CWQ) in Table~\ref{tab:time}. 
This evaluation simulates the online inference latency for answering a question using different RAG approaches. 
All experiments were conducted on the same GPU server to ensure consistent evaluation conditions.
The experimental results indicate that P-RAG reduces the time per question by 29\% to 36\% compared to Standard RAG. Notably, the Combine Both baseline, which showed the best performance and significant improvements in the main experiment, requires almost the same online computation time as the Standard RAG method.
In contrast, multi-round RAG frameworks like DRAGIN and FLARE exhibit significantly higher latency for answering a question compared to single-round methods.
For both P-RAG and Combine Both baselines, we present the inference times separately from the time required for merging and loading the LoRA (0.32s). 
This distinction arises because, in our current implementation, the time spent on merging and loading the LoRA significantly exceeds theoretical expectations. 
The floating-point operations involved in the LoRA operation step contribute less than $1\%$ to the total computational cost of generating a response~\cite{hulora}, but the latency of memory loading and data communications in our current implementation is far from perfect.
We believe this latency can potentially be addressed through engineering optimizations. 
It is important to note that our analysis primarily emphasizes the relative time ratios and trends across the different methods, as actual application times and latencies can vary depending on hardware configurations, such as CPU, GPU, memory, and storage.

\section{Conclusion and Future Directions}
\begin{table}
\caption{The average time required by the LLaMA3-8B model to answer a question on the 2WikiMultihopQA (2WQA) and ComplexWebQuestions (CWQ) datasets. The "+0.32" footnote for P-RAG and Combine Both indicates the total time needed for merging and loading the LoRA adapter.}
\label{tab:time}
\centering
\setlength\tabcolsep{3.6pt}
\begin{tabular}{llclc}
\toprule
                      & \multicolumn{2}{c}{\textbf{2WQA}}                                              & \multicolumn{2}{c}{\textbf{CWQ}}                                                \\
                      \cmidrule(lr){2-3} \cmidrule(lr){4-5} 
                      & \multicolumn{1}{c}{\textbf{Time(s)}} & \multicolumn{1}{c}{\textbf{Speed Up}} & \multicolumn{1}{c}{\textbf{Time(s)}} & \multicolumn{1}{c}{\textbf{Speed Up}}  \\
                      \toprule
                
\textbf{P-RAG}        & ${2.34}_{+0.32}$                              & 1.29x                              & ${2.07}_{+0.32}$                              & 1.36x                               \\
\textbf{Combine Both}      & ${3.08}_{+0.32}$                              & 0.98x                              & ${2.84}_{+0.32}$                             & 0.99x                               \\

\textbf{Standard RAG} & 3.03                                      & 1.00x                                 & 2.82                                      & 1.00x                                  \\
\textbf{FLARE}        & 10.14                                     & 0.25x                              & 11.31                                      & 0.25x\\
\textbf{DRAGIN}       & 14.60                                     & 0.21x                              & 16.21                                     & 0.17x\\   
\toprule
\end{tabular}
\end{table}

This work introduces Parametric RAG, a novel framework that addresses the limitations of in-context knowledge augmentation by parameterizing external documents. Parametric RAG infuses these parameterized documents directly into the model, reducing contextual overload and online computational costs while maintaining robust performance. 
Our experiments on multiple benchmarks demonstrate that Parametric RAG outperforms traditional retrieval-augmented generation methods across different LLMs. 
Ultimately, Parametric RAG offers a more efficient and scalable pathway to integrate external knowledge into LLMs, paving the way for further innovation in parametric-based knowledge augmentation.

Despite its significant potential, Parametric RAG presents several challenges that warrant further investigation. First, the current parameterization process is computationally intensive, and the parametric representations of each document are substantially larger than plain text. Future work could explore more methods to improve computational and storage efficiency, making the parameterization process more scalable. Second, the parameterized documents are currently tied to specific LLMs, restricting their ability to generalize across different models. Developing universal, model-agnostic representations could significantly enhance flexibility and reuse across diverse systems. 
Finally, we believe the potential applications of information parameterization can be extended beyond RAG. 
For instance, LLM-based agents could benefit from parameterizing the agent’s profiles and configuration, which could alleviate context-length constraints and improve online computational efficiency.
By addressing these challenges, future research could unlock more potential for the Parametric RAG paradigm.

\bibliographystyle{ACM-Reference-Format}
\bibliography{sample-base}

\end{document}